\documentclass[journal]{IEEEtran}
\usepackage{booktabs}
\usepackage{graphicx} 
\usepackage{caption}
\usepackage{amsmath}
\usepackage{diagbox}
\usepackage{cite}
\usepackage{color}
\usepackage{subcaption}
\usepackage{float}
\usepackage{multirow}
\usepackage{booktabs}
\usepackage{amsfonts}
\usepackage[backref]{hyperref} 
\hypersetup{hidelinks}

\ifCLASSINFOpdf
\else
\fi
\begin{document}

\title{
% Bare Demo of IEEEtran.cls\\ for IEEE Journals
Social Force Embedded Mixed Graph Convolutional Network for Multi-class Trajectory Prediction  
}

\author{
Quancheng Du, Xiao Wang, \textit{Senior Member}, \textit{IEEE}, 
Shouguo Yin, 

Lingxi Li, \textit{Senior Member}, \textit{IEEE}, and Huansheng Ning, \textit{Senior Member}, \textit{IEEE}
% Quancheng Du, Rui Ai, Yuxiao Wang and Weihao Gu

        % Michael~Shell,~\IEEEmembership{Member,~IEEE,}
         %Aug 31, 2023
        % \today
        % John~Doe,~\IEEEmembership{Fellow,~OSA,}
        % and~Jane~Doe,~\IEEEmembership{Life~Fellow,~IEEE}% <-this % stops a space
% \thanks{Manuscript received June 25, 2023.}
\thanks{
% This work was supported by the National Natural Science Foundation of China under Grant 62173329. (\textit{Corresponding author: Xiao Wang}.)
This work was supported by the National Natural Science Foundation of China under Grant  62173329. (\textit{Corresponding author: Xiao Wang}.)

Quancheng Du is with the School of Computer and Communication Engineering, University of Science and Technology Beijing, Beijing 100083, China. (E-mail: quancheng\textsubscript{-}du@xs.ustb.edu.cn)

Xiao Wang is with the School of Artificial Intelligence, Anhui University, Hefei 230031, China, and also with the State Key Laboratory for Management and Control of Complex Systems, Institute of Automation, Chinese Academy
of Sciences, Beijing 100190, China. (E-mail: xiao.wang@ahu.edu.cn)

Shouguo Yin is with the School of Artificial Intelligence, Anhui University, Hefei 230031, China. (E-mail: 2638711962@qq.com)

Lingxi Li is with the School of Electrical and Computer Engineering, Indiana University-Purdue University Indianapolis, Indianapolis IN46202, USA. (E-mail: ll7@iupui.edu)

Huansheng Ning is with the School of Computer and Communication Engineering, University of Science and Technology Beijing, Beijing 100083, China. (E-mail: ninghuansheng@ustb.edu.cn)

}% <-this % stops a space
% \thanks{J. Doe and J. Doe are with Anonymous University.}% <-this % stops a space
}


\markboth{Journal of \LaTeX\ Class Files,~Vol.~14, No.~8, August~2023}%
{Shell \MakeLowercase{\textit{et al.}}: Bare Demo of IEEEtran.cls for IEEE Journals}

\maketitle

\begin{abstract}

Accurate prediction of agent motion trajectories is crucial for autonomous driving, contributing to the reduction of collision risks in human-vehicle interactions and ensuring ample response time for other traffic participants.
Current research predominantly focuses on traditional deep learning methods, including convolutional neural networks (CNNs) and recurrent neural networks (RNNs).
These methods leverage relative distances to forecast the motion trajectories of a single class of agents. However, in complex traffic scenarios, the motion patterns of various types of traffic participants exhibit inherent randomness and uncertainty. Relying solely on relative distances may not adequately capture the nuanced interaction patterns between different classes of road users.
In this paper, we propose a novel multi-class trajectory prediction method named the social force embedded mixed graph convolutional network (SFEM-GCN). The primary goal is to extract social interactions among agents more accurately. 
SFEM-GCN comprises three graph topologies: the semantic graph (SG), position graph (PG), and velocity graph (VG). These graphs encode various of social force relationships among different classes of agents in complex scenes. 
Specifically, SG utilizes one-hot encoding of agent-class information to guide the construction of graph adjacency matrices based on semantic information. PG and VG create adjacency matrices to capture motion interaction relationships between different classes agents. These graph structures are then integrated into a mixed graph, where learning is conducted using a spatio-temporal graph convolutional neural network (ST-GCNN).
To further enhance prediction performance, we adopt temporal convolutional networks (TCNs) to generate the predicted trajectory with fewer parameters. 
Experimental results on publicly available datasets demonstrate that SFEM-GCN surpasses state-of-the-art methods in terms of accuracy and robustness.

\end{abstract}


\IEEEpeerreviewmaketitle

\section{Introduction}

The development of autonomous driving technology is expected to bring about profound changes in intelligent transportation systems\cite{chen2018parallel,liu2020parallel,ghorai2022state}. 
It has the potential to improve road safety, reduce traffic congestion, and provide more convenient modes of transportation\cite{tian2023vistagpt}.
Within the domain of autonomous driving technology, trajectory prediction assumes a pivotal role. It enables autonomous vehicles to navigate intricate traffic environments by analyzing and anticipating the behaviors of surrounding traffic participants.
To achieve safe and agile autonomous driving, references \cite{huang2022survey,wang2023safety} outlined requirements for trajectory prediction in terms of speed, accuracy, and generalization. However, when dealing with diverse classes of road users, such as pedestrians, cars, and cyclists, the prediction of the trajectory for autonomous driving still faces challenges due to the complexity of the scenarios and the variability of interactive behaviors\cite{shi2023trajectory}.
\begin{figure*}[t]
      \centering
      \includegraphics[width=1\textwidth]{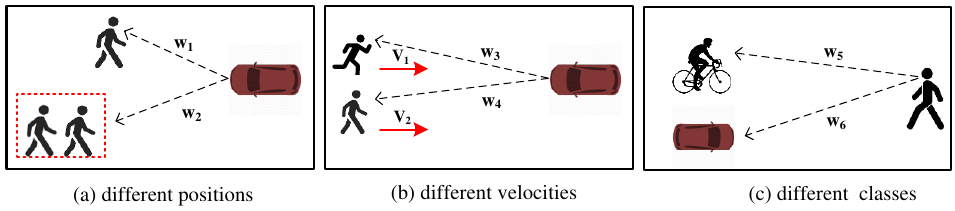}
      \caption{Three different factors that need to be considered in the scenario, with $w$ representing attention weights. }
       \label{factors}
\end{figure*}

In early research, pooling mechanisms grounded in recurrent neural networks (RNNs) exhibited promising performance in modeling social interactions.
S-LSTM \cite{alahi2016social} is a classic application of RNN to predict the trajectory of pedestrians, which captures the movements of pedestrians using long short-term memory (LSTM) and incorporates social pooling mechanisms to model interactions among pedestrians. 
This approach not only surpasses traditional rule-based trajectory prediction in terms of accuracy but also demonstrates commendable robustness in complex scenarios.
To further advance trajectory prediction research, SGAN \cite{gupta2018social} introduced generative adversarial networks (GANs) on top of S-LSTM to address the multimodal distribution of pederstrian motion and demonstrated advantages in robustness and accuracy.
Subsequent research has incorporated attention mechanisms to more effectively model the impact of various agents on the target \cite{zhang2022ai}.
SoPhie \cite{sadeghian2019sophie} added social and physical attention modules to assign different weights to surrounding pedestrians and scenes. 
Social-BiGAT \cite{kosaraju2019social} introduced a graph-based GAN that models social interactions among all pedestrians in a static scene using graph attention networks. It generates socially compatible trajectories by establishing a reversible mapping between predicted trajectories and the behavioral characteristics of the target.

In recent years, graph convolution networks (GCNs) have garnered significant attention for accurately representing topological relationships among agents and achieving competitive results\cite{mo2023predictive,sheng2022graph,golchoubian2023pedestrian}. 
SSTGCNN \cite{mohamed2020social} employed a GCN-based approach that utilizes relative distances to represent interactions between pedestrians. This method combines GCNs with temporal convolutional networks (TCNs) to effectively handle temporal dependencies.
SGCN \cite{shi2021sgcn} modeled the interaction between pedestrians as a directed graph, capturing not only the interactions that significantly affect the target but also the directional information between pedestrians.
While GCN-based methods have demonstrated some effectiveness, real-world traffic scenarios are exceptionally complex, posing a challenge to accurately model social interactions among agents using simple graph structures\cite{lv2023ssagcn}.
Drawing inspiration from the social force relationship, it is understood that the social factors influencing agents' future trajectories encompass multiple aspects, including relative position, velocity, and agent class\cite{shenyu}.
For example, in Figure \ref{factors}(a), it depicts the interaction between agents at different positions. Since closer proximity is more likely to attract attention, a higher weight should be assigned to this interaction, i.e., $w_1 > w_2$. In Figure \ref{factors}(b), assuming $V_1 > V_2$, it illustrates the interaction between agents with different velocities. Considering that higher velocity has a greater impact on the interaction, a larger weight should be assigned to the agent with a higher velocity, i.e., $w_3 > w_4$. In Figure \ref{factors}(c), it represents the interaction between agents with different classes. Compared with a cyclist, a car is more likely to capture human attention, thus requiring a higher weight to be assigned to the car, i.e., $w_6 > w_5$. 
Some methods have overlooked the aforementioned factors and primarily focused on relative distance as the basis for modeling interactions, such as through position pooling or calculating attention weights based on distance\cite{alahi2016social,mohamed2020social}. 
Additionally, these methods have predominantly addressed the prediction of trajectories for single class, lacking relevant research on predicting interactions among multi-class agents.

A recent study, Semantics-STGCNN \cite{rainbow2021semantics}, introduced a novel approach that incorporates agent label features into the adjacency matrix of the velocity representation. This integration effectively incorporates class information, leading to state-of-the-art performance in multi-class trajectory prediction.
However, this method overlooks the influence of position information on trajectory prediction, which can significantly impact accuracy in real-world traffic scenarios\cite{xue2018ss}.
In this paper, inspired by \cite{rainbow2021semantics, mohamed2020social}, we present a novel approach called social force embedded mixed graph convolutional network (SFEM-GCN) to model interaction relationships among different classes agents. SFEM-GCN incorporates three types of graph topologies: semantic graph (SG), position graph (PG), and velocity graph (VG). The goal is to achieve effective trajectory prediction by comprehensively considering the social factors that influence the movement of agents. Specifically, semantic information, position information, and velocity information of agents in the scene are encoded and processed. Social force relationships in the scene are then constructed through mixed multiple graph-guided adjacency matrices.
We adopt a multi-layer spatial temporal graph convolutional neural network (ST-GCNN) with residual structures is employed to aggregate node features on the graph, and TCNs is used to predict the trajectories of multi-class agents in real-world scenarios. 
In summary, the main contributions of this paper are as follows.

\begin{enumerate}
    \item We propose a multi-class trajectory prediction method based on social force embedded mixed graph convolutional network (SFEM-GCN) that encodes social force relationship about different social behaviors. We employ three graph topologies to delineate social interactions among agents of different classes. In contrast to existing methods, our proposed mixed graph structure captures raw information between agents and represents various social interaction features, offering a more intuitive and effective model for social interactions.
    \item By increasing the number of convolutional layers in our model, we have successfully improved trajectory prediction within the SSTGCNN \cite{mohamed2020social} architecture. This augmentation has led to noteworthy improvements in the overall performance of the model.
    % This enhancement has resulted in significant improvements in the overall performance of the model.
    \item Our method is evaluated on a widely used trajectory prediction dataset SDD, showing that SFEM-GCN outperforms existing methods.
    In particular, compared to state-of-the-art multi-class trajectory prediction methods with a GCN model proposed recently\cite{rainbow2021semantics}, our model achieves approximately 3\% improvement in the Minimum Average Displacement Error (mADE) and 4\% improvement in the Minimum Final Displacement Error (mFDE). Furthermore, in the latest metrics, Average$^{2}$ Displacement Error (aADE) and Average Final Displacement Error (aFDE), our method shows approximately 8\% and 13\% performance improvement, respectively.
\end{enumerate}

The remainder of the paper is organized into the following sections:
Section \ref{sec:related_work} introduces a brief overview of the related work.
Section \ref{sec:method} describes the problem statement and the details of our proposed method.
Section \ref{sec:expe} presents the experimental setup and results.
Finally, Section \ref{sec:conc} concludes the paper.

\section{Related Work}\label{sec:related_work}

\subsection{Modeling Social Interactions for Trajectory Prediction}
Trajectory prediction seeks to forecast the future motion paths of agents based on provided historical trajectory data, typically encompassing a brief interval of 3 to 5 seconds\cite{xu2022adaptive,zuo2023trajectory,xu2022group}. Traditional trajectory prediction has primarily made use of model-driven methods, employing complex mathematical and statistical models to characterize the motion patterns of agents\cite{yamaguchi2011you,ye2016hybrid}. 
For example, in early studies, 
Helbing \textit{et al.} \cite{helbing1995social} pioneered the concept of social force, introducing an energy potential field (repulsive and attractive forces) to model social interactions among pedestrians. Kooij \textit{et al.} \cite{kooij2014context} integrated Bayesian filters and kinematic models to formulate a context-sensitive dynamic network for predicting the trajectories of agents. Furthermore, Gaussian process \cite{ellis2009modelling} and Markov decision process \cite{kitani2012activity} have been employed to model social interactions in trajectory prediction.
In recent years, with the continuous advancement of deep learning techniques, data-driven artificial intelligence methods have emerged as a popular research direction for trajectory prediction\cite{ding2023incorporating,korbmacher2022review,huang2022survey}. 
The S-LSTM model, introduced by Alahi \textit{et al.} \cite{yamaguchi2011you}, stands as a classic application of RNN in predicting pedestrian trajectories. Leveraging LSTM, the model adeptly captures the intricate motion patterns of pedestrians while integrating a social pooling mechanism to model interaction relationships.
SGAN\cite{gupta2018social} employed GAN to predict multimodal trajectories and introduced a novel pooling mechanism to compute social interactions based on relative distances between pedestrians.
Beyond the previously mentioned networks, GCNs have also emerged as a popular approach for accurately predicting trajectories of agents within various scenes.
SSTGCNN \cite{mohamed2020social} embedded social interactions among pedestrians in an adjacency matrix, modeling them as a graph network structure. It utilized kernel functions to process the adjacency matrix and captured spatial and temporal interaction information. STGAT \cite{huang2019stgat} learned the influence weights between pedestrians through graph attention networks, resulting in outstanding pedestrian trajectory prediction.

Although previous methods have successfully modeled social interactions among agents and achieved satisfactory prediction results, they often do not fully consider the various social factors that influence agents. 
For instance, the influence of agents with different positions, velocities, and classes is crucial for trajectory prediction \cite{westny2023mtp}.
Furthermore, existing methods have primarily focused on trajectory prediction for single class agents, overlooking the influence of interaction dynamics among different classes. 
In this paper, we propose a mixed graph convolutional network that comprehensively considers the influence of the factors mentioned above, with the goal of achieving more accurate multi-class trajectory prediction. To capture complex social interactions, we construct a multi-type graph-guided adjacency matrix that represents the interaction relationships within the scene. This approach enables precise calculation of interaction degrees between agent nodes within the graph structure, resulting in a comprehensive representation of social force relations among agents and enhanced accuracy in capturing social interaction features.

\subsection{Graph Convolutional Networks in Trajectory Prediction}
GCN is a neural network capable of handling graph-structured data, extending the concept of convolutional neural networks (CNNs) to graphs\cite{wu2020comprehensive}. 
GCN applies discrete convolutions to weigh and aggregate node and edge features, assigning different weights based on neighboring nodes to effectively capture the interaction relationships\cite{teng2023motion}. Due to its powerful representation capabilities in non-Euclidean spaces, GCN has been widely applied in trajectory prediction tasks in recent years \cite{jain2016structural,mohamed2020social}.
For example, Huang \textit{et al.}\cite{huang2019stgat} introduced the STGAT model, which treats each pedestrian as a node on the graph and shares information among different pedestrians. 
Mohamed \textit{et al.} \cite{mohamed2020social} proposed a social spatial-temporal graph convolutional neural network (SSTGCNN). This model embeds social interaction relationships between pedestrians into the adjacency matrix, treating them as a graph structure. It utilizes kernel functions to process the adjacency matrix, capturing both spatial and temporal interactions.
Considering the directional nature of pedestrian interactions, Shi \textit{et al.}\cite{shi2021sgcn} introduced the sparse graph convolution network (SGCN) for pedestrian trajectory prediction, with the aim of addressing the modeling redundancy issue in existing approaches when dealing with dense undirected interactions among pedestrians. However, it's worth noting that the use of directed graphs can lead to computationally complex models that consume more computational resources \cite{liu2022stmgcn}.
In addition, the emergence of graph attention networks (GATs) \cite{mo2022multi} has yielded promising predictive performance. STGAT \cite{huang2019stgat} enhances the interaction between pedestrians by introducing a flexible spatio-temporal graph attention mechanism. Social-BiGAT \cite{kosaraju2019social} learns reliable feature representations for agent interactions using GAT, resulting in accurate multimodal trajectory prediction. GAT enhances trajectory prediction by processing encoded trajectory information and provides the model with interpretability.

Despite the significant progress of GCN-based methods in the field of trajectory prediction, they exhibit limitations in practical applications. For instance, they demand a substantial amount of data to discern differences between nodes in the graph and consume more computational resources \cite{lv2023ssagcn}. In light of this, drawing inspiration from reference \cite{rainbow2021semantics}, we propose a mixed graph convolutional network based on social force relation to address these limitations. In contrast to existing GCN methods, our approach emphasizes manual design by incorporating social force relationships and domain expertise to guide the model construction process. Consequently, our method requires fewer computational parameters and achieves faster execution speed. Additionally, by considering more social factors, our method can better differentiate the influences between different nodes, leading to more accurate prediction.

\section{Methodology}\label{sec:method}

\subsection{Problem Statement}
The trajectory prediction can be fundamentally transformed into a decision problem of time series\cite{zhang2023spatial}. Specifically, given observation of $M$ agents in the scene over a fixed duration in the past, their historical trajectories are denoted as $Y^{i} =\left\{Y^{1},Y^{2},\ldots,Y^{M}\right\}$ where $i \in \left \{ 1,2,...,M\right \}$. The observed trajectory of the $ i$-$th$ agent at time step $t$ can be expressed as 
\begin{equation}\
    Y_{t}^{i} =\left\{\left.\left(x_t^i,y_t^i\right)\in\mathcal{R}^2\right|t=1,2,\ldots, T_{\text{obs}}\right\}
\end{equation}
where $\left(x_t^i,y_t^i\right)$ denotes the true position coordinates. Over observation time-steps $1\le t\le T_{\text{obs}}$, we aim to predict the likely trajectories for the $i$-${th}$ agent in the future time-steps $T_{\text{obs}}+1\le t\le T_{\text{pred}}$, denoted as
\begin{equation}
    {\hat{Y}}_{_t}^{i} =\left\{\left({\hat{x}}_t^i,{\hat{y}}_t^i\right)\in\left.\mathcal{R}^2\right|t=T_{\text{obs}}+1,\ldots{},T_{\text{pred}}\right\}
\end{equation}

Similarly to \cite{alahi2016social,mohamed2020social}, our method employs a bivariate Gaussian distribution to characterize the coordinates $(x_t^i,y_t^i)$ of agent trajectories as random variables within the probability distribution of agent positions. We assume that the predicted agent trajectory coordinates $\hat{Y}_{_t}^{i}=(\hat{x}_t^i, \hat{y}_t^i)$ also follow a bivariate Gaussian distribution. The model parameters are learned by minimizing the negative log-likelihood loss, as shown in the following:
\begin{equation}
\tilde{Y}_t^i \sim N\left(\tilde{\mu}_t^n, \tilde{\sigma}_t^n, \tilde{\rho}_t^n\right)
\end{equation}
% \begin{equation}
% L_n\left(W_{\text {all }}\right)=-\sum_{t=T_{o b s}+1}^{T_{o b s}+T_{\text {pred }}} \log \left(\tilde{Y}_i^t \sim N\left(\tilde{\mu}_t^n, \tilde{\sigma}_t^n, \tilde{\rho}_t^n\right)\right)   
% \end{equation} 
\begin{equation}
L(W)=-\sum_{t=T_{\text {obs }}+1}^{T_{\text {pred }}} \sum_{i=1}^N \log \left(\tilde{Y}_t^i \sim N\left(\tilde{\mu}_t^n, \tilde{\sigma}_t^n, \tilde{\rho}_t^n\right)\right)   
\end{equation}
where $W$ represents the learned network parameters. Our goal is to minimize the loss value, allowing us to obtain optimal weights for the network. $\tilde{\mu}_t^n$ represents the mean, $\tilde{\sigma}_t^n$ represents the variances, and $\tilde{\rho}_t^n$ represents the correlation of the bivariate Gaussian distribution.

\begin{figure*}[t]
      \centering
      \includegraphics[width=1\textwidth]{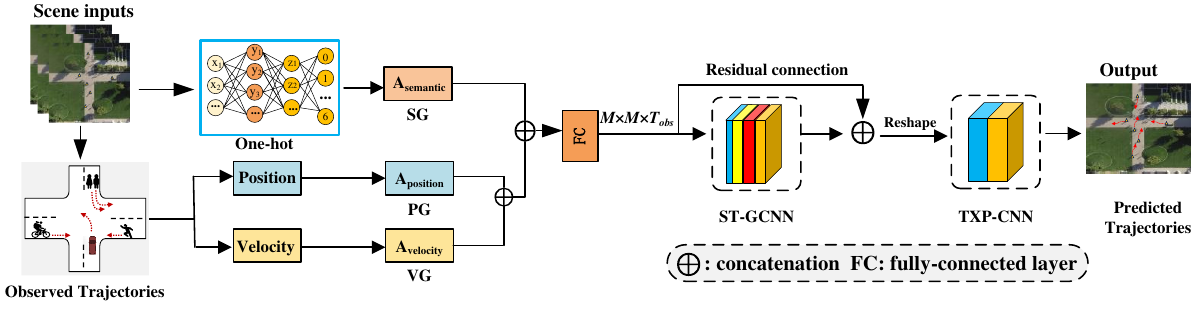}
      \caption{Overall architecture of SFEM-GCN. Our approach constructs SG, PG, and VG by embedding social force relationships. Then, we utilize ST-GCNN to aggregate the temporal and spatial feature information from the graph nodes. Finally, we employ the TXP-CNN to obtain the predicted trajectories.}
      \label{architecture}
\end{figure*}
\subsection{Overall Model}
This section present SFEM-GCN, a novel trajectory prediction model that constructs a mixed graph convolutional neural network by effectively embedding the social force relationships among multi-class agent.
The proposed method is depicted in Figure \ref{architecture}. 
The model takes both the semantic information of the scene and the historical trajectory information of the agents, which includes their position coordinates, as input. The coordinate information is then transformed into a graph representation, where each agent is depicted as a node in the scenario. Furthermore, the semantic information of the scene is encoded using one-hot encoding to generate corresponding class labels. Subsequently, we introduce a mixed graph neural network that integrates semantic, position, and velocity information to comprehensively characterize the social interaction behavior of agents within the scene.
Specifically, we have constructed three types of graph topologies and incorporated social force relationship into the adjacency matrix of the GCN, enabling accurate capture of the strength of interactions between agents. We employ ST-GCNN to extract spatio-temporal interaction features among agents in the scene. Finally, the processed graph sequence is fed into TCNs to estimate the future trajectory of the multi-class agents. 
More details of the implementation and methodology will be explained later.

\subsection{Social Force Model}
The social force model (SFM) simulates collective behavior by describing the interactions between different agents within the same environment \cite{crosato2022interaction}. This model considers both the internal motivations driving individual movements and the influence of surrounding agents on their motion. In the context of an SFM , agents are treated as point mass particles, and their movement is governed by the principles of Newton's equations of motion\cite{helbing1995social}, which are expressed as
\begin{equation}
        \frac{d^{2} \vec{p} }{dt^{2} } =\frac{d\vec{r} }{dt} =\frac{\Vec{f }_{\text{SFM}} }{m} 
\end{equation}
where $\vec{p}$ and $\Vec{r}$ are represent the agent position and velocity respectively. The mass of the agent, denoted as $m$, is a scalar value that represents the agent's inertia and responsiveness to external forces.
$\vec{f }_{\text{SFM}}$ represents the social force between agents, which influences their accelerations. It is computed by considering the interactions between an agent and its surrounding environment as well as other agents. 
This modeling approach aims to simulate motion behaviors similar to pedestrians, encompassing path selection, obstacle avoidance, and interactions with other agents. By embedding the social force relationships into the graph neural network, we have developed SFEM-GCN. This integration enables a more accurate simulation of agent behavior, leading to more realistic and detailed trajectory predictions.

\subsection{Position Graph Topology Representation}\label{position}
Building upon SFM, it is established that the shorter the relative distance between neighboring agents, the greater their mutual influence\cite{lv2023ssagcn,li2020pedestrian}. Similarly to \cite{mohamed2020social}, we use the relative positions of agents to construct a position graph topology. First, we construct a set of spatial graphs $G_{t}^{p}$ that represent the relative positions of agents in the scene at each time step $t$. $G_{t}^{p}$ is defined as $G_{t}^{p}=(V_{t},E_{t}^{p})$, where $V_{t}=\left \{p_{t}^{i}|\forall i\in \left \{ 1,...,M \right \} \right \} $ represents the set of vertices on the graph $G_{t}^{p}$. The observed positions $\left ( x_{t}^{i}, y_{t}^{i} \right )$ are the attributes of the corresponding vertices $p_{t}^{i}$. $E_{t}^{p}$ is the set of edges in the graph $G_{t}^{p}$ that is represented as $E_{t}^{p}=\left \{ e_{t}^{ij} \mid \forall i,j \in \left \{1,...,M \right \} \right \} $. If $v_{t}^{i}$ and $v_{t}^{j}$ are connected, $e_{t}^{ij}=1$; otherwise, $e_{t}^{ij}=0$. To accurately model the degree of mutual influence between neighboring agents at the time step $t$, we introduce the adjacency matrix $A_{t}^{p}$ and adopt the kernel function for the calculation as follows:
\begin{equation}
A_{t}^{p}(i,j)=
\left\{  
             \begin{array}{lr}  
             \frac{1}{\parallel p_{t}^{i} - p_{t}^{j} \parallel_{2} } ,
               & \text{if} \ p_{t}^{i} \ne p_{t}^{j}\\  
              0,
             & \text{otherwise} \\  
             \end{array}  
\right.
\end{equation}
where $i$ and $j$ are the indices of two agents in the observed scene; $p_{t}^{i}=  ( x_{t}^{i}, y_{t}^{i}  )$ is the position of agent $i$ at time step $t$; $ p_{t}^{j}= ( x_{t}^{j}, y_{t}^{j}  )$ is the position of agent $j$ at time step $t$; $\parallel. \parallel_{2}$ is the $l_{2}$ norm. Considering that agents at longer distances have a diminished influence on trajectories, we employ the inverse Euclidean distance to define the edge set of the position graph.

\subsection{Velocity Graph Topology Representation}

Velocity information plays a pivotal role in modeling interactions among agents \cite{yang2020pedestrian}. While position-based interaction accurately characterizes social interactions through relative distances, the velocity of each agent is a significant factor influencing the future trajectories of others in real-world scenarios \cite{su2022trajectory}. In practical settings, faster-moving agents often attract more attention from their surroundings due to rapid movement, leading to quicker approaches and an increased risk of collisions \cite{minoura2022utilizing, zhou2023acp}. To comprehensively consider agents' velocity information and enhance the accuracy of predicting future interaction behaviors, we construct a velocity-based graph network as the second sub-graph to model relationships between agents.

Similarly to the construction of the graph network based on position in \ref{position}, we have also built a velocity graph, denoted by $G_{t}^{v} =\left ( V_{t},E_{t}^{v}  \right ) $ to model interactions relationship between agents. Here, $V_{t}=\left \{v_{t}^{i}| \forall i\in \left \{ 1,...,M \right \} \right \} $ is the set of vertices with $M$ nodes, and $E_{t}^{v}$ represents the set of edges modeling the geometric correlations between nodes in the velocity graph $G_{t}^{v}$, defined by the $M\times M$ adjacency matrix $A_{t}^{v}$, which is expressed as
\begin{equation}
A_{t}^{v}(i,j)=
\left\{  
             \begin{array}{lr}  
             \frac{1}{\parallel v_{t}^{i} - v_{t}^{j} \parallel_{2} } ,
               & \text{if} \ v_{t}^{i} \ne v_{t}^{j}\\  
              0,
             & \text{otherwise} \\  
             \end{array}  
\right.
\end{equation}
where $v_{t}^{i}=\left ( x_{t}^{i}-x_{t-1}^{i} , y_{t}^{i}-y_{t-1}^{i} \right ) $ represents the moving velocity of agent $i$ at step $t$; $v_{t}^{j}=\left ( x_{t}^{j}-x_{t-1}^{j} , y_{t}^{j}-y_{t-1}^{j} \right ) $ represents the moving velocity of agent $j$ at step $t$.

\subsection{Semantic Graph Topology Representation}
In this section, we introduce the construction of the third sub-graph, namely the semantic graph. It involves creating an adjacency matrix guided by the label information to connect different classes of agents, such as cars, pedestrians, and cyclists. Considering that different classes of agents have varying impacts on surrounding agents, in complex traffic scenarios, we may prioritize our attention towards larger vehicles approaching us rather than cyclists\cite{AAS-CN-2022-0820}. 
Therefore, prediction models based on a single class trajectory cannot fully capture real-life scenarios involving multi-class agents interacting with each other over long distances. However, incorporating label information from agents of different classes can significantly influence the relationships between trajectories. By assigning different weights to distinct agents, we can effectively address this issue.

Inspired by \cite{rainbow2021semantics}, we initially employ one-hot encoding to convert different classes of agents as vectors, where each vector has length $L$ and depends on the number of classes of agents in the scene. From the one-hot encoding, the corresponding classes of agents are encoded as 1, while the remaining positions are set to 0\cite{wang2023safety}. Through $M$ iterations (assuming $M$ agents in the scene), we create a tensor $T$ of size $L\times M$. Subsequently, we construct graph relationships based on label information and establish a label-based adjacency matrix between objects. Specifically, we replicate the rows and columns of both $T$ and the transpose matrix $T^{'}$ in the list using one-hot encoding, resulting in two tensors of size $ M \times M$, each of dimension $L$. These tensors are then reshaped and concatenated along the label dimension $L$, producing a tensor $C$ with the size of $M\times M \times 2L$, which represents the interactive relationships between agents. Finally, a fully connected layer is employed to extract pertinent characteristics from the label information in $C$, generating a semantic adjacency matrix $A^{s} $ with trainable parameters. $A^{s} $ automatically encodes valuable features from label information, acquiring information crucial for trajectory prediction.

We fusion the position graph, the velocity graph, and the semantic graph to construct a mixed graph network, then connect $A_{t}^{p} $, $A_{t}^{v} $, and $A^{s} $ to the characteristic dimension, and the resulting adjacency matrix is passed through a fully connected layer to produce an adjacency matrix $A^{f}$ with a size of $M\times M\times T_\text{obs}$. The fusion process can be formalized as 
\begin{equation}\label{fc}
 A^{f}=f(A_{t}^{p},A_{t}^{v} ,A^{s} ) 
\end{equation}
where $f$ denotes the fully connected layer.

\subsection{The Spatial-Temporal Graph Convolution Neural Network}
Following the construction of the graph topology, we employ ST-GCNN to integrate the graph representation. This integration is achieved through the application of a spatio-temporal convolution process, as detailed in \cite{rainbow2021semantics}. By applying a kernel, this operation extends the array-to-graph convolution process to aggregate features within the local neighborhood of a graph across both spatio-temporal dimensions. Furthermore, we normalize the adjacency matrix $A^f$ at each time step $t$, following the methodology of\cite{mohamed2020social}, ensuring a precise extraction of the features. The expression is as follows:
\begin{equation}
    \hat{A} _{t}^{f} =\Lambda _{t}^{-\frac{1}{2}} \left ( A_{t}^{f}+I \right ) \Lambda _{t}^{-\frac{1}{2}} 
\end{equation}
where $I$ is the identity matrix and $\Lambda _{t}$ represents the diagonal node degree matrix with each element corresponding to the row summation of the matrix $A_{t}^{f}+I$. $A_t$ refers to the adjacency matrix at time $t$, which has been normalized to incorporate weights. $\hat{A} _{t}^{f} $ is the normalized weighted adjacency matrix at time $t$, ensuring positive semi-definite for spectral decomposition in the GCN. The ST-GCNN is further defined by convoluting the position graph $P(l)$, the velocity graph $V(l)$ and the semantic graph $S(l)$, with the kernel $W (l)$ in layer $l$ under the graph laplacian $\hat{A} _{t}^{f}$. The computational process is formulated as 
\begin{equation}
    \varphi \left ( V(l) ,A\right )=\delta \left ( \Lambda  _{t}^{-\frac{1}{2} } \hat{A} _{t}^{f}\Lambda _{t}^{-\frac{1}{2}} V(l)W(l) \right )  
\end{equation}

\subsection{Trajectory Generation}
Beyond considering the spatial information of agents, incorporating the temporal information of trajectories is crucial for accurate trajectory prediction. Drawing inspiration from \cite{mohamed2020social}, we employ a novel approach that combines TCN and CNN techniques to process multi-class agent trajectories, named the time extrapolator convolutional neural network (TXP-CNN). By leveraging convolutional operations in the temporal domain, we effectively capture intricate temporal relationships across multiple time scales. This enables us to make sequential predictions while addressing the issue of error accumulation commonly observed in RNN-based methods. Specifically, we employ a temporal stacking approach in which multiple convolutional layers are stacked along the temporal dimension of the output feature map $\hat{V}$ originating from the last ST-GCNN layer. Each output from the temporal layer is then residual connected to the previous layer.

\begin{table*}[t]
  
  \centering
  \caption{\MakeUppercase{Quantitative comparisons with mADE and mFDE (minimum error).}}
  \label{tab:transposed}
  \renewcommand{\arraystretch}{1.2}%row space 
    \setlength\tabcolsep{4pt}
  \footnotesize
  
  % \resizebox{2\columnwidth}{!}{    
  \begin{tabular}{c||cccccccccc}
    \toprule
     \diagbox{\textbf{Metrics}}{\textbf{Methods}}&\textbf{Linear} & \textbf{SF} & \textbf{S-LSTM} & \textbf{SGAN}&  \textbf{CAR-Net}& \textbf{DESIRE} &\textbf{Trajectron++}& \textbf{SSTGCNN} & \textbf{Semantics-STGCNN}&\textbf{Ours} \\
      \midrule
    \textbf{mADE  $\downarrow$} & 37.11 & 36.48  & 31.19 & 27.25 & 25.72 & 19.25&19.30 & 26.46 & 18.12 &  \textbf{15.31}\\
    \textbf{mFDE  $\downarrow$}   & 63.51 & 58.14 & 56.97 & 41.44 & 51.80 & 34.05 &32.70& 42.71 &29.70&\textbf{25.72}\\
    \bottomrule
  \end{tabular}
  \label{tab:compar}
% }
\end{table*}
\section{EXPERIMENT AND EVALUATION}\label{sec:expe}
We conduct evaluation experiments on a widely used benchmark dataset and compare the results with other state-of-the-art methods. Specifically, we evaluate SFEM-GCN model on the stanford drone dataset (SDD) \cite{chiara2022goal}, which offers trajectory data and class labels for various object classes. The SDD comprises six different classes: biker, pedestrian, car, cart, bus, and skater. The scenes in the SDD are captured from a top-down view using drones, and the tracking coordinates are measured in pixels.
The structure of GCN used in SFEM-GCN is similar to those\cite{alahi2016social,rainbow2021semantics,lv2023ssagcn}, we also input the trajectory of eight time steps (3.2 s) and predict the next 12 time steps (4.8 s). To assess the effectiveness of our method, we randomly sampled 20 trajectories ($K$ = 20) from the predicted multinomial distribution. 

\subsection{Experiments }
\begin{enumerate}    
    \item Evaluation Criteria: Following the approach employed by other baseline, the performance of the model is evaluated using the Average Displacement Error (ADE) and the Final Displacement Error (FDE) metrics introduced in \cite{mohamed2020social}. ADE measures the average prediction error across all time steps, while FDE focuses on the error at the final time step. Recognizing the multimodal characteristics of agent motion, we leverage the bivariate Gaussian distribution generated by the proposed model and adopt the Minimum Average Displacement Error (mADE) and Minimum Final Displacement Error (mFDE) metrics\cite{xu2022groupnet} to evaluate the performance of our model. To do so, we sample $K$ instances from the predicted distribution and identify the instances that produce the minimum errors. mADE and mFDE are used as the standard metrics, which are defined as 
\begin{equation}
\begin{aligned}
 mADE_k=\frac{\sum_{i=1}^{N} \min _{k=1}^{K} \sum_{t=T_{\text{obs}+1}}^{T_{\text {pred }}}\left\|Y_t^i-\widehat{Y}_t^{i k}\right\|_2}{{~N} \times\left(T_{\text {pred }}-T_{\text{obs}}\right)}
\end{aligned}
\end{equation}

\begin{equation}
\begin{aligned}
 mFDE_k=\frac{\sum_{i=1}^{{N}} \min _{k=1}^{{K}}\left\|{Y}_t^i-\widehat{{Y}}_t^{{ik}}\right\|_2}{{~N}}, {t}=T_{\text {pred }}  
\end{aligned}
\end{equation}
 where the ground truth of the future trajectories is denoted as $Y_{t}^{i} =(x_{t}^{i},y_{t}^{i})$, and $\hat{Y}_{t}^{ik}$ represents the predicted position of the $k$-$th$ ($1\le k\le K$) sampled trajectory at time $t$.
Recognizing that relying solely on minimum errors may not adequately demonstrate the robustness of the generated bivariate Gaussian distributions, we introduce new metrics referred to as the Average$^{2}$ Displacement Error (aADE) and the Average Final Displacement Error (aFDE)\cite{rainbow2021semantics}. These metrics offer a more comprehensive comparison of models and are defined as follows:
\begin{equation}
 \begin{aligned}
{aADE}=\frac{\sum_{i=1}^{{N}} \sum_{t=T_{\text {obs }+1}}^{T_{\text {pred }}} \sum_{{s}=1}^{{S}}\left\|{Y}_t^i-\widehat{{Y}}_t^{i {s}}\right\|_2}{{~N} \times T_{\text {pred }} \times {S}}
\end{aligned}
\end{equation}

\begin{equation}
\begin{aligned}
{aFDE}=\frac{\sum_{i=1}^{{N}} \sum_{{s}=1}^{{S}}\left\|{Y}_t^i-\widehat{{Y}}_t^{i s}\right\|_2}{{~N} \times {S}}, {t}=T_{\text {pred }}   
\end{aligned}
\end{equation}
where $S$ denotes a set of samples generated by sampling from a bivariate Gaussian distribution. The metrics aADE and aFDE are more accurate than mADE and mFDE, as they take into account $S$ samples from the predicted bivariate Gaussian distribution and average the errors across these samples.
    \item Implementation Details: 
    % The structure of SFEM-GCN used is similar to \cite{mohamed2020social, rainbow2021semantics,lv2023ssagcn}. 
    The proposed solution was implemented using the PyTorch deep learning framework\cite{paszke2019pytorch}. The NVIDIA GV100 GPU was used for training and evaluation. The model was trained for 250 epochs with a batch size of 128. Stochastic gradient descent (SGD) was employed as the neural network optimizer. The initial learning rate was set to 0.01, and the decay was set to 0.002 after 150 epochs. The PReLU activation function \cite{he2015delving} was utilized in the model architecture. 
    % According to our ablation study in Table ???\ref{tab:s4}, the best model to use has four ST-GCNN layers and two TXP-CNN layers.
    
\end{enumerate}

% and aADE, and aFDE (average error)

\subsection{Baselines}
We compared the proposed method with state-of-the-art baselines as follows:
\begin{enumerate}  
   \item Linear\cite{alahi2016social}: This method utilizes a linear regressionor to estimate the linear parameters by minimizing the least squares error. 
   \item SF\cite{yamaguchi2011you}: This method incorporates the implementation of the social force model, which includes modeling several factors such as group affinity and predicted destinations.
    \item S-LSTM\cite{alahi2016social}: The proposed method is an LSTM-based trajectory prediction model that incorporates an innovative pooling mechanism.
    \item  DESIRE \cite{lee2017desire}: This method introduces a conditional variational auto-encoder system, which is based on the concept of inverse optimal control.
    \item SGAN \cite{gupta2018social}: This method introduces a GAN into pedestrian trajectory prediction and incorporates global pooling for interaction modeling.
    \item CAR-Net \cite{sadeghian2018car}: This method introduces an attentive recurrent model called CAR-Net, which integrates the information of saliency regions within the scenes.
    \item Trajectron++ \cite{salzmann2020trajectron++}: This method has the ability to integrate diverse data sources beyond historical trajectory information and is capable of generating predictions for future conditions that adhere to dynamical constraints. It can be classified as a generative multi-agent trajectory prediction method.
    \item SSTGCNN \cite{mohamed2020social}: This method models pedestrian trajectories as graphs and utilizes STGCNN to capture and analyze social interactions.
     \item Semantics-STGCNN\cite{rainbow2021semantics}: This method introduces a state-of-the-art model for multi-class trajectory prediction based on an undirected graph architecture. Leveraging undirected edges in the graph, this model effectively captures complex relationships and dependencies between different classes of objects, leading to accurate trajectory predictions.
   
\end{enumerate}
\subsection{ Quantitative Analysis}
\begin{table}[t]
  \centering
  \caption{\MakeUppercase{Quantitative analysis with aADE and aFDE (average error).}}
  \label{tab:transposed}
  \renewcommand{\arraystretch}{1.2}%row space 
    \setlength\tabcolsep{5pt}  % \setlength\tabcolsep{}调节横向宽度，\tiny用来调节字体大小
  \footnotesize
     
  \begin{tabular}{c||cc||c}
    \toprule
     \diagbox{\textbf{Metrics}}{\textbf{Methods}}& \textbf{SSTGCNN} & \textbf{Semantics-STGCNN}&\textbf{Ours} \\
      \midrule
    \textbf{aADE  $\downarrow$} & 49.96 & 33.14 & \textbf{25.45}\\
    \textbf{aFDE  $\downarrow$} & 90.10 & 61.14 &\textbf{48.28}\\
    \bottomrule
  \end{tabular}
  \label{tab:s2}
\end{table}

In this subsection, we present a comparison between our proposed method and nine recently proposed methods. Table \ref{tab:compar} shows the results achieved by each method in various scenarios. 
Based on these findings, our method has demonstrated outstanding performance in multi-class trajectory prediction. Specifically, the Linear method \cite{alahi2016social}, which assumes a linear correlation between the motion attributes of agents, tends to deviate significantly from real-world social patterns, resulting in comparatively less accurate results. SF \cite{yamaguchi2011you} provides a simplistic approach to trajectory prediction but has limitations in capturing the complexity and individuality of human behavior in real-world scenarios. Due to the inherent errors introduced during trajectory prediction by RNN, LSTM-based methods \cite{alahi2016social, gupta2018social} have failed to achieve competitive results. This also highlights the limitations of the pooling structure based on LSTM hidden states in accurately extracting interaction features in complex scenarios. Although CAR-Net \cite{sadeghian2018car} can incorporate salient information from the scene, its performance remains suboptimal due to the uncertainty of agent movements and the complexity of the environment. DESIRE \cite{lee2017desire}, which uses a conditional variational auto-encoder model for multimodal trajectory prediction, achieves promising results. Trajectron++ \cite{salzmann2020trajectron++}, with its advanced prediction framework and unique modeling approach, consistently remains at the forefront of trajectory prediction methods, outperforming contemporaneous approaches in terms of evaluation criteria.
Our method builds on the foundations of SSTGCNN\cite{mohamed2020social} and Semantics-STGCNN\cite{rainbow2021semantics}, which fully incorporates prior knowledge of social interactions between different classes of agents, enabling the model to better balance socice force relationships among agents.
Taking into account the significant contributions mentioned above, our model achieves a 11\% reduction in mADE and a 17\% reduction in mFDE compared to SSTGCNN. Compared to Semantics-STGCNN, our method reduces mADE by 3\% and mFDE by 4\%.
\begin{table}[t]
  \centering
  \caption{\MakeUppercase{Network comparisons with the number of trainable parameters and inference time (s).}}
  \label{tab:transposed}
  \renewcommand{\arraystretch}{1.2}%row space 
    \setlength\tabcolsep{5pt}
  % \resizebox{1\columnwidth}{!}{    
  \footnotesize
  \begin{tabular}{c||cc||c}
    \toprule
     \diagbox{\textbf{Metrics}}{\textbf{Methods}}& \textbf{SSTGCNN} & \textbf{Semantics-STGCNN}&\textbf{Ours} \\
      \midrule
    \textbf{Parameters Number  $\downarrow$} & 7563 & 7852 & \textbf{4420}\\
    \textbf{Inference Time  $\downarrow$} & 0.302 & 0.309 &\textbf{ 0.276}\\
    \bottomrule
  \end{tabular}
  \label{tab:s3}
% }
\end{table}

In Table \ref{tab:s2}, we conducted a comprehensive comparison of the experimental results for the evaluation metrics aADE and aFDE, as introduced in a recent method \cite{mohamed2020social,rainbow2021semantics}. Our method takes into account the influence of position, velocity, and semantic information on social interactions among agents, leading to state-of-the-art performance in comparative experiments. Specifically, we observed significant improvement by 8\% and 13\% in the aADE and aFDE metrics, respectively.

Furthermore, we place significant emphasis on the inference time of our method, recognizing its crucial role in ensuring the safety of autonomous driving systems. Table \ref{tab:s3} provides a detailed comparison of the model's parameter count and inference time. Encouragingly, our method, designed for agent interaction modeling, not only reduces the parameter count but also achieves faster inference speed, making it the optimal choice. This also highlights the advantages of social force-based interaction models.
Importantly, even with the utilization of multiple graphs to capture social force relationships among different classes agents, the inference performance of our method remains unaffected.

\subsection{Qualitative Analysis}
\begin{figure}[t]
      \centering
      \includegraphics[width=0.5\textwidth]{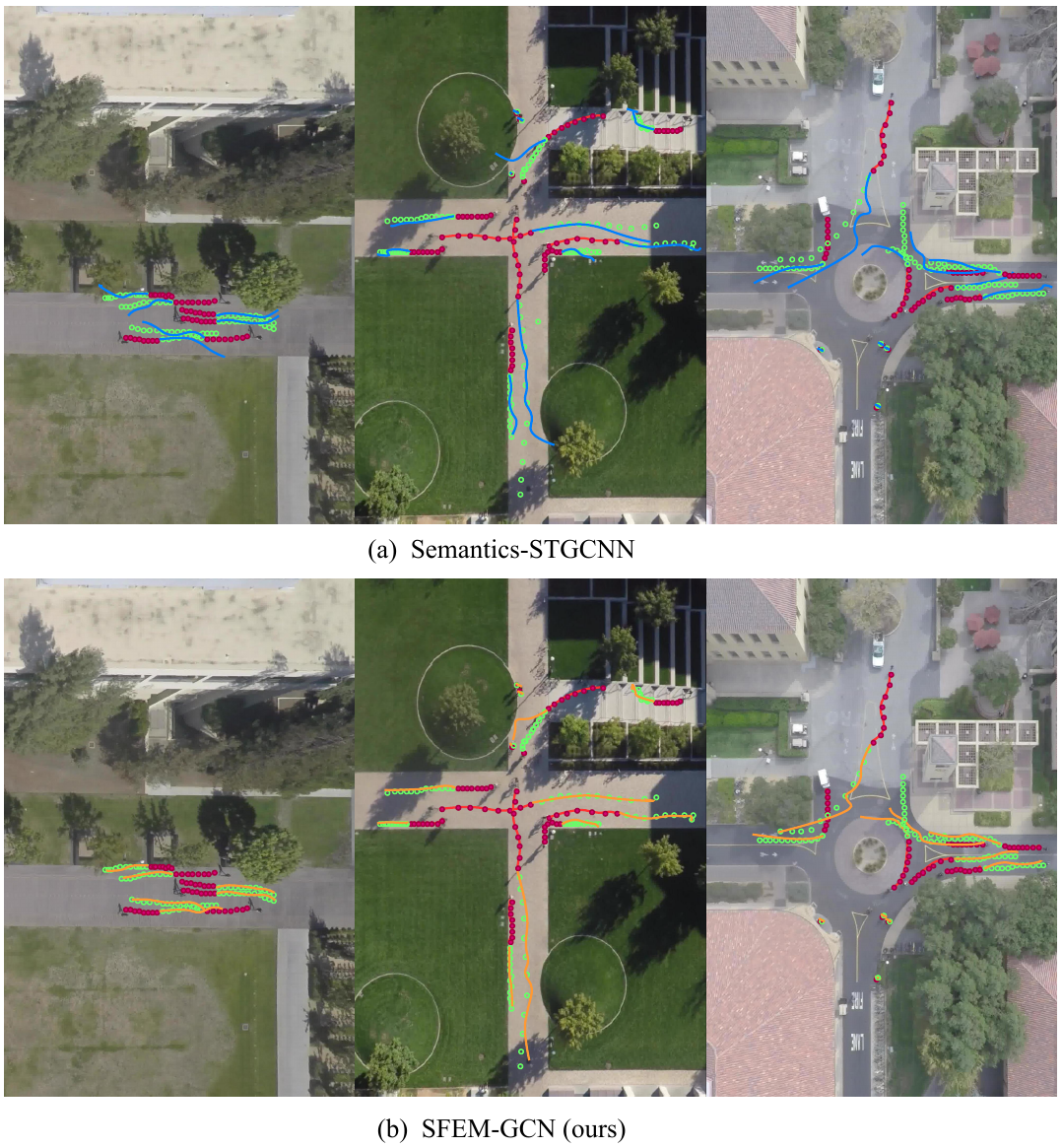}
      \caption{Visualization of the trajectory distribution predicted by compare models of multi-class agent motion on SDD. Red filled circle is observed trajectories (8 frames), green hollow circles are ground-truth (12 frames), blue lines in (a) are predicted results by Semantics-STGCNN, orange lines in (b) are predicted results by the proposed SFEM-GCN.}
      \label{visual}
\end{figure}

In Figure. \ref{visual}, we showcase the experimental prediction results of two methods, Semantics-STGCNN and SFEM-GCN, for one frame from three different scenarios of the SDD. Overall, our model outperforms Semantics-STGCNN by generating more realistic predicted trajectories. Specifically, in the leftmost image in Figure. \ref{visual}, where there is minimal interaction among the agents, even when the motion trajectories between agents exhibit a linear relationship, Semantics-STGCNN fails to handle the behaviors of agents at different positions effectively, resulting in deviations between the predicted and ground truth. As for the middle image in Figure. \ref{visual}, where most agents move in different directions and significant interaction occurs, Semantics-STGCNN shows noticeable directional biases, leading to significant deviations between the predicted and ground truth trajectories, particularly in destination predictions. For example, in the result of Semantics-STGCNN from
the second subfigure on the middle-hand side, the cyclist's predicted destination as the lawn is evidently unreasonable. In contrast, our predictions align closely with the ground truth. Additionally, in the rightmost image in Figure. \ref{visual}, we present the motion of multiple agent classes in a crowded scene, which resembles real-world autonomous driving scenarios and poses more complex challenges in handling. From the visual results, both Semantics-STGCNN and SFEM-GCN exhibit some incorrect predictions. For example, in the circular road scenario depicted in the image, both methods make erroneous predictions regarding the agents' turning intentions. Considering the higher uncertainty in predicting the intentions of different agents in crowded and highly dynamic scenes, such erroneous predictions are understandable. However, compared to Semantics-STGCNN, SFEM-GCN achieves better trajectory prediction results in most cases by incorporating position and velocity graph to balance the social force relationships among agents and utilizing semantic graph to assign different label as guidance for various agents, enabling more effective learning of the differences between trajectories of multi-class agents. In summary, the visual results from these three different scenarios demonstrate that the three constructed graph topologies effectively simulate the social interactions between multi-class agents, leading to more realistic trajectory prediction.
\subsection{Ablation study}
\begin{enumerate}
    \item  \textbf{Ablation study on model architecture:}
   Our method is constructed by stacking multiple layers of ST-GCNN and TXP-CNN. The selection of appropriate parameters for the network structure is crucial to optimizing the results of trajectory prediction. Therefore, we employ a similar approach as presented in \cite{chai2020multiobjective} to test and select the optimal parameters for the network structure. In each experiment, we maintained the same hyperparameter settings for both the testing and training stages. The maximum number of layers for ST-GCNN and TXP-CNN was set to 5. The experimental results can be found in Table \ref{tab:s4}. Through testing with different numbers of network layers, we found that setting ST-GCNN and TXP-CNN to 4 layers and 2 layers, respectively, yielded optimal values for the metrics.

% 消融实验1，不同的神经网络层数
\begin{table}[t]
  \centering
  \caption{\MakeUppercase{Ablation study of our model. The first row corresponds to the number of TXP-CNN layers, and the first list
corresponds to the number of ST-GCNN layers.}}
  \label{tab:transposed}
  \renewcommand{\arraystretch}{1.2}%row space 
    \setlength\tabcolsep{2pt}
  % \scriptsize
  \footnotesize
  % \resizebox{1\columnwidth}{!}{ 
  \begin{tabular}{c||ccccc}
    \toprule
    \footnotesize 
     \textbf{Methods}& \textbf{1} & \textbf{2}&\textbf{3}  & \textbf{4}&\textbf{5}\\
      \midrule
      % \midrule
    \textbf{1} & 18.35/31.16 & 19.27/34.61  & 18.89/32.98 & 20.72/37.34 & 20.25/32.76 \\
    \textbf{2} & 16.85/29.11 & 20.61/30.33  & 17.29/28.65 & 18.88/33.28 & 21.25/34.71 \\
    \textbf{3} & 20.67/36.94 & 15.95/27.88  & 21.15/37.38 & 20.84/29.71 & 18.16/31.50 \\
    \textbf{4} & 17.68/31.22 & \textbf{15.31/25.72}  & 21.75/38.64 & 17.48/27.81 & 17.50/31.72 \\
    \textbf{5} & 24.51/34.91 & 20.28/33.95  & 25.84/45.93 & 20.11/31.06 & 19.99/33.61 \\
 
    \bottomrule
  \end{tabular}
  \label{tab:s4}
% }
\end{table}
% 消融实验2，不同的网络结构
\begin{table}[t]
  \centering
  \caption{\MakeUppercase{Ablation study of our model in terms of mADE/mFDE/aADE/aFDE metrics.}}
  % \renewcommand{\arraystretch}{1.2}%row space 
  % \resizebox{1.0\columnwidth}{!}{   
  \renewcommand{\arraystretch}{1.2}%row space 
  % \resizebox{0.5\columnwidth}{!}{
    \setlength\tabcolsep{13pt}
    % \tiny
    \footnotesize   
    
  \begin{tabular}{c||cccc}
    \toprule

     \textbf{Methods}& \textbf{mADE} & \textbf{mFDE} &\textbf{aADE}  & \textbf{aFDE}\\
      \midrule
    \textbf{w/o SG} & 17.46 & 28.68 & 33.04 & 61.36 \\
    \textbf{w/o PG} & 26.65 & 52.09 & 39.11 & 78.16  \\
    \textbf{w/o VG} & 19.88 & 28.25 & 29.36 & 56.58 \\
    \textbf{Ours} & \textbf{15.31} & \textbf{25.72}  & \textbf{25.45} &\textbf{48.28}\\
    
    \bottomrule
  \end{tabular}
  \label{tab:s5}
% }
\end{table}
    
    \item  \textbf{Ablation study on model components:}
   To demonstrate the performance of each component in the proposed model, we tested three different variants on the SDD. The test results are presented in Table \ref{tab:s5}. Specifically, ``w/o SG" indicates the removal of the semantic information sub-graph component, ``w/o PG" represents the removal of the position information sub-graph component, and ``w/o VG" represents the exclusion of the velocity information sub-graph component. 
   The results indicate that each component of the proposed model significantly contributes to the accuracy of trajectory prediction. Among these, the positional information between agents has a more pronounced impact on trajectory prediction.
   Specifically, the adoption of multi-graph fusion enables effective modeling of social interaction relationships between different classes of agents. Additionally, guidance based on social force relationship is crucial for improving the performance of the model.

\end{enumerate}

\section{Conclusion}\label{sec:conc}
In this paper, we propose SFEM-GCN, a multi-class trajectory prediction method that incorporates various factors of social and scene interaction to construct a mixed graph topology based on social interaction. 
% The goal is to achieve accurate and reasonable trajectory predictions. 
When handling social interaction information between different classes of agents, we use coordinate data to calculate the positions and velocities of the agents, constructing corresponding adjacency matrices to measure their influences. Simultaneously, we consider scene context information by constructing a graph adjacency matrix guided by semantic information, highlighting the varying influences of different classes in the scene. Attention weights are computed by incorporating information from the label. The constructed mixed graph topology is then input to ST-GCNN with residual structures, and predicted trajectories are generated using TCNs. We conducted experiments on the publicly available dataset SSD and compared our method with state-of-the-art approaches. The quantitative analysis of the experimental results has shown that our method outperforms existing methods in terms of accuracy and generalization ability of trajectory prediction. Furthermore, qualitative analysis confirms that our method accurately predicts the motion trajectories of multi-class agents in the scene, yielding more plausible trajectory predictions.

While our method has demonstrated promising performance on the open dataset SSD, accurately predicting the future motion trajectories of agents in high-dynamic and crowded scenes remains a challenging task. For instance, issues such as inaccurate inference of agent turning intentions leading to prediction failures need to be addressed. Therefore, further research is necessary in addressing these challenging scenarios. Additionally, future work can explore the use of attention mechanisms \cite{zhang2022trajectory, yang2022predicting, wang2023s} in deep learning to prioritize key agents, capture their motion patterns more accurately, and achieve precise trajectory predictions.

% or


% \appendices



% \section*{Acknowledgment}


\ifCLASSOPTIONcaptionsoff
  \newpage
\fi



%\bibliographystyle{IEEEtran}


% \end{thebibliography}

\bibliographystyle{IEEEtran}
\bibliography{bibtex/bib/HAOMO_Sup}
% \bibliography{bibtex/bib/IEEEabrv}

% % \begin{IEEEbiography}[{\includegraphics[width=1in,height=1.25in,clip,keepaspectratio]{mshell}}]{Michael Shell}
% % or if you just want to reserve a space for a photo:

% % Biography text here.
% \end{IEEEbiography}
\begin{IEEEbiography}[{\includegraphics[width=1in,height=1.25in,clip,keepaspectratio]{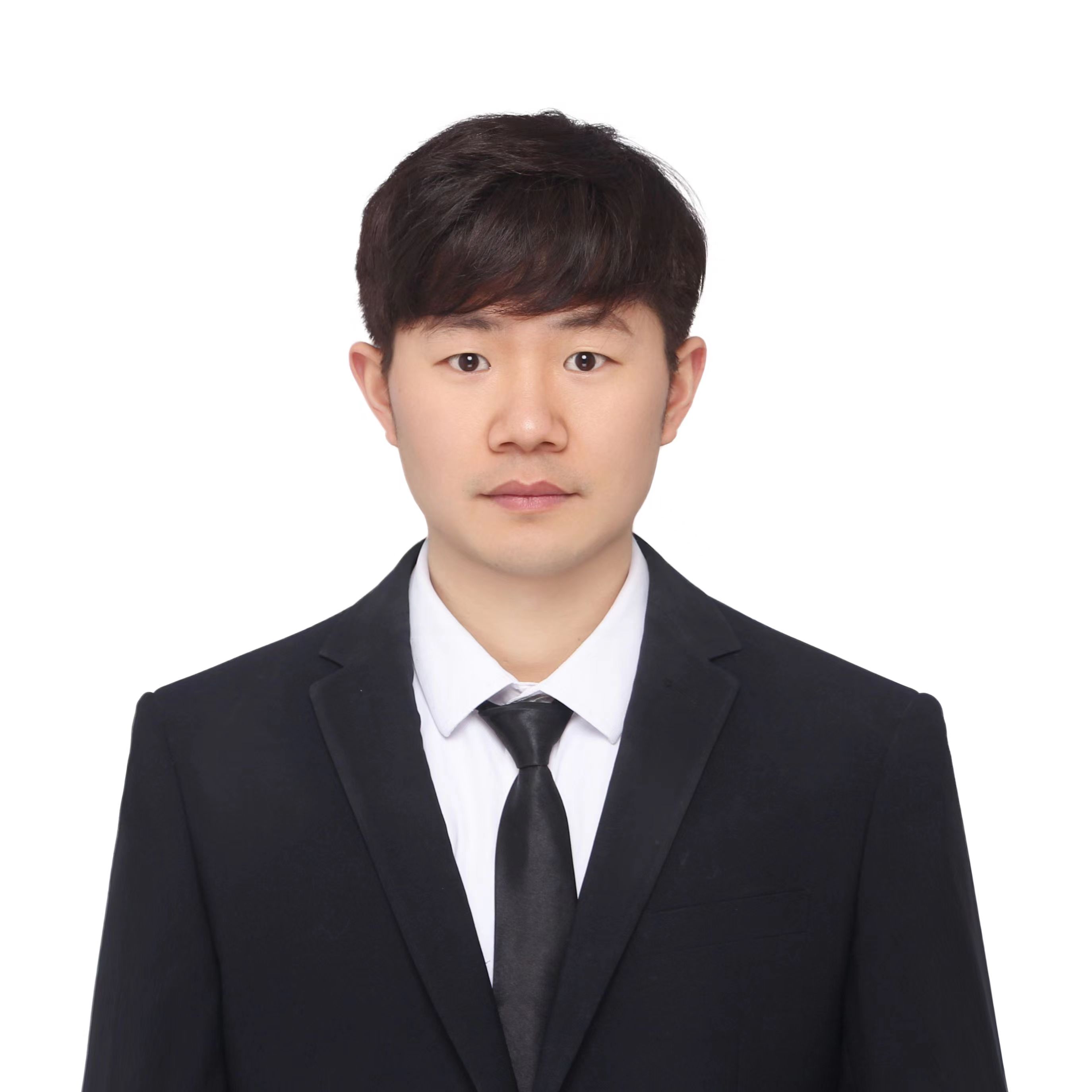}}]{Quancheng Du}
 received his B.S. degree from Henan Normal University in 2018. He currently working toward his Ph.D. degree in the School of Computer and Communication Engineering, University of Science and Technology Beijing, China. His main research interests include autonomous driving and  trajectory prediction.
\end{IEEEbiography}
\begin{IEEEbiography}[{\includegraphics[width=1in,height=1.25in,clip,keepaspectratio]{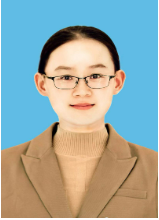}}]{Xiao Wang}
 (Senior Member, IEEE) received the Bachelor's degree in network engineering from Dalian University of Technology, Dalian, China, in 2011, and the Ph.D. degree in social computing from the University of Chinese Academy of Sciences, Beijing, China, in 2016. She is currently a professor with the School of Artificial Intelligence, Anhui University, Hefei, and the President of the Qingdao Academy of Intelligent Industries, Qingdao, Chi-na. Her research interests include social network analysis, social transportation, cybermovement organizations, and multi-agent modeling. 
\end{IEEEbiography}
\begin{IEEEbiography}[{\includegraphics[width=1in,height=1.25in,clip,keepaspectratio]{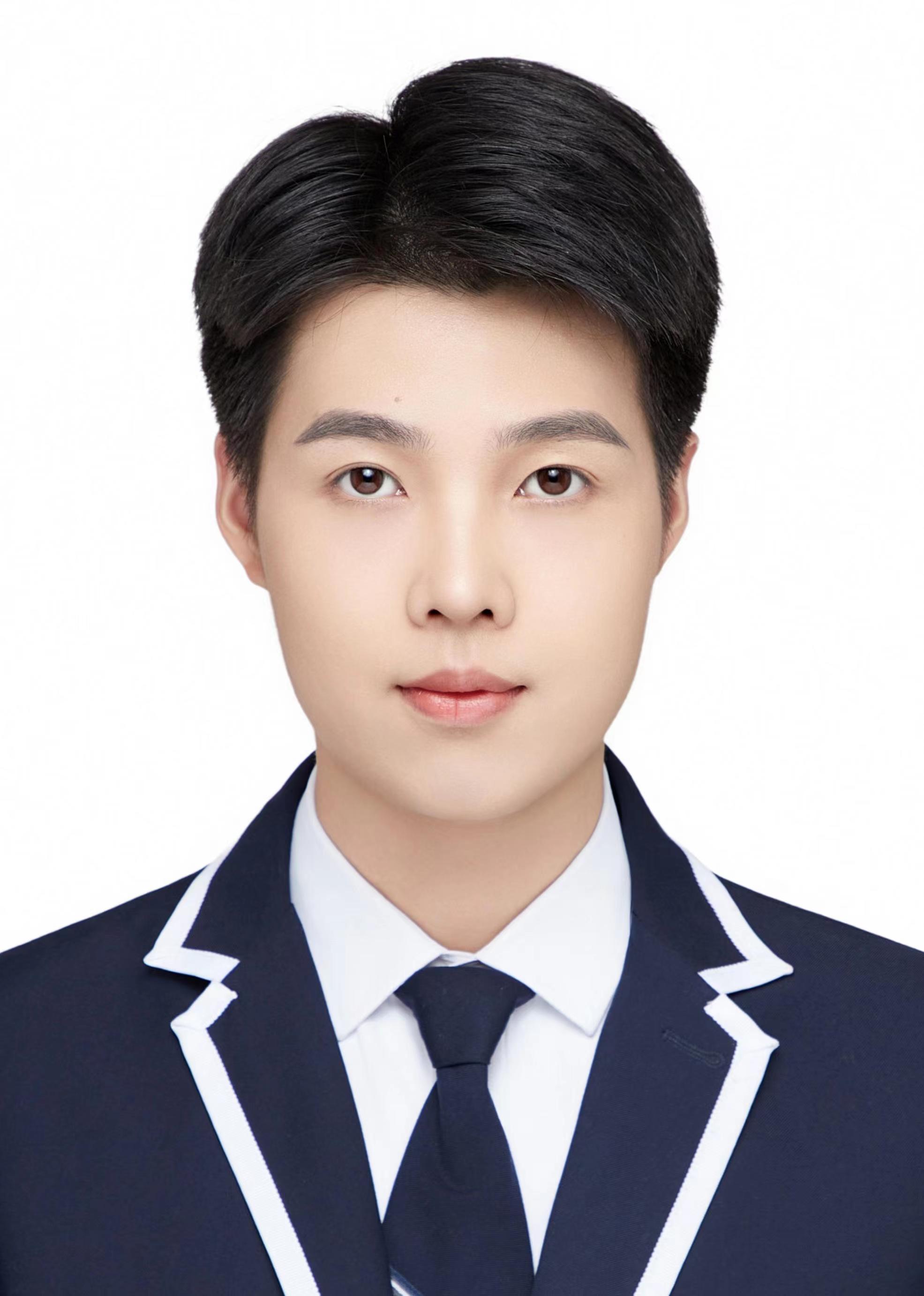}}]{Shouguo Yin}
 received a bachelor's degree from Sichuan Normal University in 2023. He is currently pursuing a master's degree at the School of Artificial Intelligence, Anhui University, China. His current research focuses on intelligent transportation systems and trajectory prediction.
\end{IEEEbiography}
\begin{IEEEbiography}[{\includegraphics[width=1in,height=1.25in,clip,keepaspectratio]{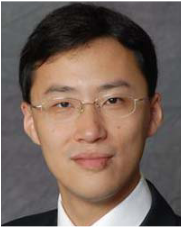}}]{Lingxi Li}
 (Senior Member, IEEE) received the B.E. degree in automation from Tsinghua University, Beijing, China, in 2000, the M.S. degree in control theory and control engineering from the Institute of Automation, Chinese Academy of Sciences, Beijing, in 2003, and the Ph.D. degree in electrical and computer engineering from the University of Illinois at Urbana–Champaign, Champaign, IL, USA, in 2008. Since August 2008, he has been with Indiana University–Purdue University Indianapolis (IUPUI), Indianapolis, IN, USA, where he is currently a Professor of electrical and computer engineering. He has authored/coauthored one book and more than 100 research articles in refereed journals and conferences. His current research focuses on the modeling, analysis, control, and optimization of complex systems, intelligent transportation systems, digital twins and parallel intelligence, and human-machine interaction.
\end{IEEEbiography}
\begin{IEEEbiography}[{\includegraphics[width=1in,height=1.25in,clip,keepaspectratio]{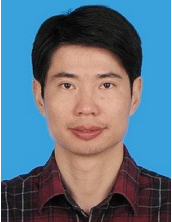}}]{Huansheng Ning}
 (Senior Member, IEEE) received his B.S. degree from Anhui University in 1996 and his Ph.D. degree from Beihang University in 2001. Now, he is a professor of the School of Computer and Communication Engineering, University of Science and Technology Beijing, China. His current research focuses on the Internet of Things and general cyberspace. He is the founder of the cyberspace and cybermatics international science and technology cooperation base.
\end{IEEEbiography}

\end{document}